\documentclass[lettersize,journal]{IEEEtran}
\usepackage{amsmath,amsfonts}
\usepackage{algorithm}
\usepackage{array}
\usepackage[caption=false,font=normalsize,labelfont=sf,textfont=sf]{subfig}
\usepackage{textcomp}
\usepackage{stfloats}
\usepackage{url}
\usepackage{verbatim}
\usepackage{graphicx}
\usepackage{cite}
\usepackage{amsthm}
\usepackage{hyperref}       
\usepackage{booktabs}       
\usepackage{nicefrac}       
\usepackage{microtype}      
\usepackage{xcolor}         
\usepackage{multirow}
\usepackage{algpseudocode}
\usepackage{amssymb}
\usepackage{caption}
\usepackage{subfig}
\usepackage{lineno}
\newtheorem{theorem}{Theorem}

\begin{document}

\title{Partial Distribution Alignment via \\ Adaptive Optimal Transport}

\author{{Pei Yang,
	    Qi Tan,
        Guihua Wen}
\thanks{Pei Yang (Corresponding Author) is with South China University of Technology, Guangzhou, China, E-mail: yangpei@scut.edu.cn.}
\thanks{Qi Tan is with South China Normal University, Guangzhou, China, Email: tanqi@scnu.edu.cn.}
\thanks{Guihua Wen is with South China University of Technology, Guangzhou, China, E-mail: crghwen@scut.edu.cn.}}

\maketitle

\begin{abstract}
To remedy the drawbacks of full-mass or fixed-mass constraints in classical optimal transport, we propose adaptive optimal transport which is distinctive from the classical optimal transport in its ability of adaptive-mass preserving. It aims to answer the mathematical problem of how to transport the probability mass adaptively between  probability distributions, which is a fundamental topic in various areas of artificial intelligence. Adaptive optimal transport is able to transfer mass adaptively in the light of the intrinsic structure of the problem itself. The theoretical results shed light on the adaptive mechanism of mass transportation. Furthermore, we instantiate the adaptive optimal transport in machine learning application to align source and target distributions partially and adaptively by respecting the ubiquity of noises, outliers, and distribution shifts in the data. The experiment results on the domain adaptation benchmarks show that the proposed method significantly outperforms the state-of-the-art algorithms. 
\end{abstract}


\section{Introduction}
\IEEEPARstart{T}{he} Optimal Transport  (OT) theory becomes a powerful tool for artificial intelligence due to its capacity to compare non-parametric probability distributions by exploiting the geometry of the underlying metric space. To name a few, optimal transport plays a crucial role in a wide variety of machine learning applications, such as generative adversarial networks \cite{ArjovskyCB17}, computer vision \cite{Solomon2016EntropicMA}, natural language processing \cite{XuZGZL20}, clustering \cite{HoNYBHP17}, semi-supervised learning \cite{Chapel2020PartialOT}, and domain adaptation \cite{CourtyFTR17}. The essential problem in these applications is how to compare two probability distributions such as aligning the fake images with the real images, aligning images with audio, or aligning the AI generated content with human feedback in large language model. 

Optimal transport \cite{MongeOT1781} aims to find an optimal way to move a pile of sand into a hole, assuming the pile and the hole must have the same volume. Optimal transport is formulated as the mathematical problem of comparing two probability distributions, which is a fundamental problem in a variety of domains. Despite its powerful capability for distribution alignment and indispensable roles in applications, a major bottleneck of classical optimal transport \cite{Villani2003TopicsIO} is that it requires the two distributions to have the same total probability mass and all probability mass has to be transported. Open-world machine learning applications exhibit the ubiquity of noises, outliers, and divergences in the data where the source and target domains usually do not follow the independent and identically distributed assumption. The classical optimal transport \cite{KantorovitchOnTT} with full-mass conservation is likely to fit noise and outliers, or undesired pairs, and prevent any form of partial matching. Caffarelli and McCaan \cite{Caffarelli2010FreeBI} and Figalli \cite{Figalli2010TheOP} proposed partial optimal transport (POT) to preserve the fixed amount of mass instead the full mass, providing flexibility for partial distribution matching. However, since there is usually no prior knowledge on the relatedness of domains,  it is a challenge on how to determine the fixed budget of mass to transport for partial optimal transport. Thus, it remains an open issue on how to align the distributions partially and adaptively. 

To this end, we propose \textit{adaptive optimal transport (AOT)} to enrich the family of optimal transport. The distinctive advantage of adaptive optimal transport against its classical counterparts lies in the ability of adaptive-mass preserving. Adaptive optimal transport determines the transported masses adaptively in the light of the intrinsic structure of the problem itself. It provides a powerful tool for partial distribution alignment by respecting the ubiquity of noises, outliers, and distribution shifts. As an instantiation application, we propose a novel machine learning paradigm based on adaptive optimal transport. It conducts the partial distribution alignment between source and target domains by treating the noises, outliers, and distribution shifts in a principled way. Furthermore, we investigate the mass allocation mechanism of adaptive optimal transport and derive the duality theory. The theoretical analysis provides insights into adaptive optimal transport and reinforces its mathematical foundation. We believe that adaptive optimal transport is of great interests to the broad areas such as artificial intelligence, biomedical, physics, operations research, urban science, etc. The main contributions of the paper are highlighted as follows.
\begin{itemize}
\item We propose adaptive optimal transport which is a novel member in the family of optimal transport. The adaptive optimal transport is distinctive from classical optimal transport in its ability of adaptive-mass preserving. 
\item The theoretical analysis regarding the mechanism of adaptive mass allocation and duality theory sheds light on the intrinsic structures of the adaptive optimal transport problem. 
\item We propose a novel machine learning paradigm based on adaptive optimal transport. It accomplishes partial distribution alignment between source and target domains. The experiments on unsupervised domain adaptation benchmarks demonstrate its effectiveness. 
\end{itemize}

Next we review the related work in Section \ref{sec2}. The formulation of adaptive optimal transport is proposed in Section \ref{sec3}, followed by the theoretical analysis in Section \ref{sec4}. The experiment results are shown in Section \ref{sec5}. Section \ref{sec6} concludes the work.

\section{Related Work}\label{sec2}
We review the related work in optimal transport and its applications to machine learning.

\subsection{Optimal Transport}
The optimal transport (OT) problem first came up in Monge's seminal work \cite{MongeOT1781}, which can be informally described as moving a pile of sand into a hole with the smallest cost. The optimal transport distance entails a rich geometric structure on the space of probability distribution. OT is formulated as the mathematical problem of comparing two probability distributions, which is of interest to many domains. Therefore, OT has become a classical subject in mathematics, probability theory, economics, optimization, etc. One of the major breakthroughs following Monge’s work was by Kantorovich \cite{KantorovitchOnTT} who was the founder of linear programming. His research in optimal resource allocation, which earned him his Nobel Prize, led him to study optimal coupling and duality, giving OT a firm footing in optimization. Many researchers in different areas found that optimal transport was strongly linked to their subjects, and helped expand the optimal transport foundations \cite{Villani2003TopicsIO}. Recent years have witnessed another revolution in the spread of OT, thanks to the emergence of approximate algorithms that can solve large-scale problems \cite{Cuturi13}. As a consequence, OT is being increasingly used to unlock various problems in artificial intelligence, statistics, bioinformatics, economics, logistics, physics, etc.  

There have been two main directions including partial optimal transport and unbalanced optimal transport to attempt to remove the constraints of full-mass preservation. Partial optimal transport \cite{Caffarelli2010FreeBI,Figalli2010TheOP} relaxed the full-mass constraints in Kantorovich's problem and preserved the fixed amount of mass. Unbalanced optimal transport \cite{Liero2017OptimalEP} relaxed the `hard' marginal constraints with the `soft' penalties by using some divergence measures. 
Robust optimal transport \cite{MukherjeeGSSY21,NietertGC22,LeNNPBH21,BalajiCF20} is similar to unbalanced optimal transport in using the soft marginal constraints measured by $f$-divergence. However, robust optimal transport emphasizes on handling the probability distributions possibly corrupted by outliers. Some other kind of robust optimal transport \cite{PratikROT2021,PatyC19} aims at maximizing the minimal transport cost over a set of parameterized ground cost functions.

Both partial optimal transport and unbalanced optimal transport (as well as robust optimal transport) provided flexibility to model partial matching to some extent. However, they lack the ability of adaptive-mass transport. For partial optimal transport, it is challenging to determine the fixed budget of mass to transport. For unbalanced optimal transport, it is usually unknown to what extent the `soft' penalties should be imposed on the marginal constraints. Therefore, we propose adaptive optimal transport in the hope of filling the gap in this field.

\subsection{Machine Learning via Optimal Transport}
Recently, optimal transport has been successfully employed in a wide variety of machine learning branches, such as generative adversarial networks \cite{ArjovskyCB17}, computer vision \cite{Solomon2016EntropicMA}, natural language processing \cite{XuZGZL20}, graph matching \cite{ZhangL24,SunYSXDCXW24}, semi-supervised learning \cite{Chapel2020PartialOT}, few-shot learning \cite{TianICCV2023}, and domain adaptation \cite{CourtyFTR17}. Also, optimal transport plays the key roles in diverse applications such as screening cell-cell communication \cite{Cangcot2023}, predicting cell responses to treatments \cite{Bunne0C22}, learning single-cell perturbation responses \cite{Bunnenot2023}.

Domain adaptation is the critical task in real-world machine learning applications since distribution discrepancy is ubiquitous in the data. Most existing works on domain adaptation can be roughly classified into two categories: discrepancy-based methods and adversarial-learning based methods. The discrepancy-based methods explicitly minimized the domain distance using discrepancy metrics such as optimal transport distance or Maximum Mean Discrepancy (MMD) \cite{BorgwardtGRKSS06}. The domain adaptation methods based on optimal transport include OTDA~\cite{CourtyFTR17}, ROT~\cite{BalajiCF20}, DeepJDOT~\cite{damodaran2018deepjdot},  JUMBOT~\cite{fatras2021unbalanced}, m-POT \cite{NguyenNV0H22}, etc. The typical methods based on MMD are JAN~\cite{LongZ0J17}, WDAN~\cite{yan2017mind}, CCD~\cite{Kang0YH19}, DeepONet~\cite{Goswami2022DeepTO}, to name a few. The adversarial-learning based methods aim to learn domain-invariant representations via adversarial training. The typical methods include DANN~\cite{ganin2016domain}, CDAN~\cite{long2018conditional}, BSP~\cite{ChenWLW19},  ALDA~\cite{chen2020adversarial}, DrugBAN~\cite{Bai2022InterpretableBA}, etc. Please refer to the survey paper \cite{WilsonC20} for more details.  

We take a closer look at some typical domain adaption methods which will be used as baselines in experiments. DEEPJDOT~\cite{damodaran2018deepjdot} is the deep learning-based extension of JDOT~\cite{courty2017joint} which is a joint distribution optimal transport method. The robust optimal transport model ROT~\cite{BalajiCF20} followed the unbalanced optimal transport formulation while keeping the $f$-divergence relaxations of marginal distributions as inequality constraints. JUMBOT~\cite{fatras2021unbalanced} adopted the unbalanced optimal transport to alleviate the issue of undesired matching during the mini-batch sampling. In contrast, m-POT~\cite{NguyenNV0H22} used partial optimal transport to mitigate the misspecified mappings by limiting the amount of masses. Domain Adversarial Neural Network (DANN) \cite{ganin2016domain} adversarially learned a feature extractor and a domain discriminator. Conditional Domain Adversarial Network (CDAN)~\cite{long2018conditional} utilized a conditional domain discriminator instead. ALDA~\cite{chen2020adversarial} combined self-training and adversarial training for noise-correction domain discrimination. The contrastive-learning based method CaCo~\cite{huang2022category} adopted the category contrastive loss for adaptation.  

The OT-based methods depend on the classical optimal transport such as Kantorovich optimal transport \cite{KantorovitchOnTT} or partial optimal transport \cite{Figalli2010TheOP} for distribution alignment. As mentioned before, they will also suffer from the limitations of full-mass or fixed-mass constraints. As the new family member of optimal transport, adaptive optimal transport is distinctive in adaptive-mass preservation, allowing for partial distribution alignment.

\section{Adaptive Optimal Transport}\label{sec3}
We propose the formulation of adaptive optimal transport, and its application in machine learning.

\subsection{The Primary Problem}
\noindent\textbf{Notation.} 
Suppose $X, Z \subset \mathbb{R}^d$ are domains in Euclidean space. $\mathcal{P}(X)$ denotes the set of nonnegative Borel measures on a space $X$. Let $\mu \in \mathcal{P}(X)$ and $\nu \in \mathcal{P}(Z)$ be two Borel measures. The density functions are $d \mu = f(x) dx$ and $d \nu = g(z) dz$.  Whenever $A$ is the Borel subset of $X$, $\mu[A]$ denotes the mass located inside $A$. We denote the space of bounded continuous functions in $X$ by $C_b(X)$. By definition, the support of a measure $\mu$ on $X$ will be the smallest closed set $A \subset X$ with $\mu[X \backslash A] = 0$, and will be denoted by $spt \mu$. Let $c(x,z)$ be the lower bounded continuous cost function which tells how much it costs to move one unit of mass from location $x \in \mathcal{X}$ to location $z \in \mathcal{Z}$. We model the transport plans by nonnegative Borel measure $\gamma \in \mathcal{P}(X \times Z)$, where $d \gamma(x,z)$ measures the amount of mass transferred from location $x$ to location $z$.

\vspace{2mm}
\noindent\textbf{The Primary Problem.}
Given two nonnegative Borel measures $\mu \in \mathcal{P}(X)$ and $\nu \in \mathcal{P}(Z)$ on the source $X$ and target $Z$ respectively, as well as the lower bounded continuous cost function $c(x,z)$, adaptive optimal transport is to find the optimal transport plans $\gamma \in \mathcal{P}(X \times Z)$ which moves the mass \textit{adaptively} from the source to the target at minimal cost. Mathematically, the adaptive optimal transport problem is formulated as
\begin{equation}\label{pp}
\mathop{\min}\limits_{\gamma \in \Gamma_{\le}(\mu,\nu)} \int_{X \times Z} c(x, z) d \gamma(x,z)
\end{equation}
where the set of admissible transport plans is denoted by $\Gamma_{\le}(\mu,\nu)$ whose left and right marginals are dominated by $\mu$ and $\nu$ respectively, i.e.
\begin{equation}\label{ppc}
\Gamma_{\le}(\mu,\nu) = \Bigg\{ \gamma \in \mathcal{P}(X \times Z) \hspace{1mm} \bigg|
\begin{array}{l}
\gamma[A \times Z] \le \mu[A] \\  
\gamma[X \times B] \le \nu[B] 
\end{array}
\Bigg\}
\end{equation}
for all Borel subset $A \subset X$ and $B \subset Z$. 
Notice that the marginal inequality constraints are used in $\Gamma_{\le}(\mu,\nu)$. Therefore, when $\mu \in \mathcal{P}(X)$ and $\nu \in \mathcal{P}(Z)$ are probability measures ($\mu[X]=\nu[Z]=1$), the transport plan $\gamma \in \mathcal{P}(X \times Z)$ is not necessarily to be a probability measure, and we have $\gamma[X \times Z] \le 1$ in this case. Also, we assume that the cost function is mixed-sign.

To achieve the goal of adaptive optimal transport, we make two relaxations from the classical optimal transport theory. First, we relax the full-mass constraints and the fixed-mass constraints required in Kantorovich optimal transport \cite{KantorovitchOnTT} and partial optimal transport \cite{Caffarelli2010FreeBI,Figalli2010TheOP}, respectively. We propose adaptive-mass preserving instead. Second, we relax the non-negative constraint on the cost function, and instead require it to be mixed-sign. The classical optimal transport usually assumes that the ground cost is non-negative \cite{Villani2003TopicsIO}. However, in many scenarios, it is naturally to allow for negative costs. 
For example, in the fields of economics and operations research, it is common to see the co-existence of both positive and negative costs. 
Consider the example related to CO$_2$ abatement from the climate change context \cite{LEVIHN20161155}. Some investment options result in financial expenses, hence the costs are positive. On the contrary, some other options both increase productivity and reduce CO$_2$  emissions, leading to financial returns. Therefore, the costs for these options are negative. 
Furthermore, from the mathematical perspective, the extension to negative costs enlarges the scope of the optimal transport problem, which could bring potential impacts to many areas.

The distinctive characteristic of adaptive optimal transport is adaptive-mass preserving. Unlike partial optimal transport or unbalanced optimal transport, it does not need to specify the fixed budget of mass or the softness of marginal constraints, which are challenging in essence. 
Adaptive optimal transport relies on both the marginal inequality constraints and the mixed-sign cost function to achieve adaptive-mass transport.
The adaptive optimal transport is capable of preserving the suitable masses in accordance with the native structures of the problem. The mass will be transferred between the active regions, while there is no allocation of mass in inactive regions. It provides an elegant solution for partial distribution matching. We will go deeper into the mass allocation mechanism of the adaptive optimal transport problem in the theoretical analysis section. Also, the by-product of adaptive optimal transport is the optimal mass transported under the optimal transport plan, which can be used as a metric to measure the relatedness of the source and the target.

\subsection{Partial Distribution Alignment}
Next, we take domain adaptation as an application area of adaptive optimal transport. In real applications, the training and test data usually do not follow the independent and identically distributed assumption. Domain adaptation aims to estimate a transferable model for target domain by exploiting source domain data in the presence of domain shift. Due to the distribution shift, the classical optimal transport with full-mass conservation is likely to fit dissimilar pairs (and noise or outliers) between source and target domains. Also, since it is unknown to what extent the two domains are related, partial optimal transport with fixed-mass conservation is restrictive. 

In the context of unsupervised domain adaption, no label is available in the target domain. Assume that $x$ and $z$ are data samples drawn from the source domain $X$ and the target domain $Z$ with uniform probability distributions $\mu$ and $\nu$ respectively. The true and predicted class probability vectors for a data sample $x$ are denoted by $p(x)$ and $q(x)$ respectively. Let $\log(\cdot)$ be a Matlab-like Logarithmic function, and $p^T(x)$ the transpose of a vector $p(x)$.

We propose a novel machine learning paradigm based on adaptive optimal transport (AOT). The objective is to minimize the adaptive optimal transport distance between the source distribution $\mu$ and the target distribution $\nu$, as well as the empirical classification loss on the source domain:
\begin{equation}\label{ppml}
\begin{aligned}
\mathop{\min}\limits_{\gamma \in \Gamma_{\le}(\mu,\nu)} & \int_{X \times Z} \Big[\alpha \big\|x - z\big\|_2^2 -  \beta p^T(x) \cdot q(z) \Big] d \gamma(x,z) \\
& - \int_X p^T(x) \cdot \log q(x) dx 
\end{aligned}
\end{equation}
where $\alpha$ and $\beta$ are non-negative coefficients. Here we use cross-entropy loss as the empirical classification loss. 

The underlying idea in constructing the cost function is to align the domains in feature space and label space simultaneously. The intuition is that the more similar the sample pair is in the both feature space and label space, the more mass transported between them. Considering only the feature space or the label space could be one-sided to define the cost function. Since the target labels are unknown, we use the  surrogate version $q(z)$. 

The entropy-regularized optimal transport \cite{Cuturi13} has the advantages that it defines a strongly convex problem which can be solved efficiently. Likewise, one may add the entropy-regularized term $\epsilon \mathcal{H}(\gamma) = - \epsilon \int_{X \times Z} \log \gamma(x,z) d \gamma(x,z)$ to the adaptive optimal transport defined in Equation \ref {ppml}, where $\epsilon$ is the entropic coefficient. The entropy-regularized term encourages the sparsity of the transport plan, and allows using the Sinkhorn-Knopp algorithm \cite{Cuturi13} for efficient computation. 

The strength of the novel machine learning paradigm is its capability of partial distribution alignment empowered by adaptive optimal transport. The noises, outliers, and distribution shifts are ubiquitous in open-world machine learning applications. Adaptive optimal transport provides a principled way for partial distribution alignment by treating the noises, outliers, and distribution shifts deliberately. Therefore, adaptive optimal transport is widely applicable to a variety of applications beyond artificial intelligence areas.

\section{Theoretical Analysis}\label{sec4}
In this section, we conduct the theoretical analysis to provide insights into the adaptive optimal transport problem. 

\subsection{AOT vs POT}
We discuss the relation and difference between adaptive optimal transport (AOT) and partial optimal transport (POT), and illustrate how AOT achieves adaptive-mass transport. 

Without loss generalization, let's assume that both $\mu$ and $\nu$ are probability measures ($\mu[X]=\nu[Z]=1$) here for simplicity. Partial optimal transport \cite{Caffarelli2010FreeBI} is formulated as
\begin{equation}
\mathop{\min}\limits_{\gamma \in \Gamma_{\le}(\mu,\nu), \atop \gamma[X \times Z] = m} \int_{X \times Z} c^+(x, z) d \gamma(x,z)
\end{equation}
where the cost function $c^+(x, z)$ is non-negative. Caffarelli and McCann introduced a Lagrange multiplier $\lambda_m \ge 0$ conjugate to the fixed-mass constraint $\gamma[X \times Z] = m$ and reformulated the POT problem as
\begin{equation}\label{potre}
\mathop{\min}\limits_{\gamma \in \Gamma_{\le}(\mu,\nu)} \int_{X \times Z} \big[ c^+(x, z) - \lambda_m \big] d \gamma(x,z).
\end{equation}
Likewise, adaptive optimal transport defined in Equation \ref{pp} can be reformulated as 
\begin{equation}\label{aotre}
\mathop{\min}\limits_{\gamma \in \Gamma_{\le}(\mu,\nu)} \int_{X \times Z} \Big[ \big(c(x, z) + \lambda_c \big) - \lambda_c \Big] d \gamma(x,z)
\end{equation}
where the cost function is mixed-sign and $\lambda_c = \mathop{\max}_{x,z}[-c(x,z)]$. The reformulation provides insights into the mechanism of adaptive-mass transport in AOT. However, adaptive optimal transport is essentially different with partial optimal transport.

First, the fundamental difference is that AOT preserves adaptive-mass while POT transports fixed-mass. For the POT problem, the goal of introducing the Lagrange multiplier $\lambda_m$ is to remove the  fixed-mass constraint $\gamma[X \times Z] = m$, making it easier to solve the POT problem. However, this does not eliminate the limitation that it needs to specify the mass budget $m$ (or equivalently find the appropriate value of the Lagrange multiplier $\lambda_m$), which is challenging because we usually have no prior knowledge on how much mass should be transported. For AOT, we have no such a fixed-mass constraint, thus there is no need to introduce an extra Lagrange multiplier. 

Second, AOT determines the mass according to the task structure, while POT relies on the user to specify the mass budget. The reformulation gives some insights into how AOT attains adaptive-mass transport. According to \cite{Caffarelli2010FreeBI}, for each mass $m$ there is a unique $\lambda$ corresponding to the $m$, and $m$ increases continuously as $\lambda$ is increased. For POT, the specific value of the Lagrange multiplier $\lambda_m$ is irrelevant to the task structure itself. On the contrary, AOT self-determines the total mass, relying on the native structure the ground costs.  Specifically, a definite $\lambda_c$ in AOT results in a definite mass, while a larger $\lambda_c$ leads to the more mass to be transported. 

Last but not least, AOT provides a much larger capacity than POT by exploring the whole spectrum of mass instead of the fixed-mass. Consider the optimal transport problems with parameterized cost functions \cite{PratikROT2021,PatyC19}. Denote the parameterized cost function by $c_{\theta}(x,z)$ where $\theta$ is the learnable parameter. The adaptive optimal transport with parameterized cost function can be formulated as 
\begin{equation}
\mathop{\min}\limits_{\theta, \atop \gamma \in \Gamma_{\le}(\mu,\nu)} \int_{X \times Z} c_{\theta}(x, z) d \gamma(x,z)
\end{equation}
Likewise, it can be reformulated as 
\begin{equation}
\mathop{\min}\limits_{\theta, \atop \gamma \in \Gamma_{\le}(\mu,\nu)} \int_{X \times Z} \Big[ \big(c_{\theta}(x, z) + \lambda_{c_{\theta}} \big) - \lambda_{c_{\theta}} \Big] d \gamma(x,z)
\end{equation}
where  $\lambda_{c_{\theta}} = \max_{x,z}[-c_{\theta}(x,z)]$. The total transport mass of AOT could increase continuously from 0 to 1 as $\lambda_{c_{\theta}}$ increases. Therefore, AOT could attain the adaptive-mass ranged continuously across the whole spectrum of mass, thus offering a much larger capacity to search for the learnable parameters. In contrast, POT sticks to the user-specified mass budget no matter how the cost functions are varying, which is likely to be trapped into local optimums.

In summary, the distinctive advantages of AOT against POT lie in three aspects: a) adaptive-mass preserving, b) self-determining according to task structure, c) larger capacity by exploring the spectrum of mass.  Since the classical optimal transport with full mass constraints can be viewed the special case of partial optimal transport by setting $m=1$, the claims hold for the classical optimal transport too.

\subsection{Adaptive Mass Transport}
We first exploit the mass allocation mechanism of the adaptive optimal transport problem. Theorem \ref{mass} reveals the relations between cost function and mass allocation, and suggests that the prerequisite of mass transportation between the sample pair is that it has a non-positive cost. Theorem \ref{ar} indicates that the masses are transferred between active regions only.

\vspace{2mm}
\begin{theorem}[\textbf{Optimal Transport Mass}]\label{mass}
Let $\gamma^*$ be the optimizer for the adaptive optimal transport problem defined in Equation \ref{pp}. For the pair $(x,z)$ with positive cost, there is no mass transferred between them. For the pair $(x,z)$ with negative cost, either the mass taken from $x$ coincides with $d \mu(x)$, or the mass transferred to $z$ coincides with $d \nu(z)$. That is to say,  \\ 
(i) If $c(x,z) > 0$, then $d\gamma^*(x,z) = 0$;  \\
(ii) If $c(x,z) < 0$, then at least one equation holds:
\begin{equation}
\int_Z d \gamma^*(x,z) = d \mu(x)
\end{equation}
\begin{equation}
\int_X d \gamma^*(x,z) = d \nu(z)
\end{equation}
And it is not necessary that both equations hold.
\end{theorem}

\vspace{2mm}
\begin{proof}
Proof by contradiction.  

\vspace{1mm}
\noindent (i) We assume $d \gamma^*(x,z) > 0$ if $c(x,z) > 0$. Let's set $\bigtriangleup m = d \gamma^*(x,z)$.
By letting $d \gamma^*(x,z) = 0$ which still satisfies the partial mass constraints defined in Equation \ref{ppc}, the objective of the adaptive optimal transport problem in Equation \ref{pp} will decrease by $c(x,z) \cdot \bigtriangleup m$. In this way, we obtain a solution which is better than $\gamma^*$. This contradicts with the premise that $\gamma^*$ is the optimizer for the adaptive optimal transport problem. Hence, it arrives $d \gamma^*(x,z) = 0$.

\vspace{1mm}
\noindent (ii) Assume that both equations do not hold for $c(x,z) < 0$, i.e.,
\begin{equation}\nonumber
\int_Z d \gamma^*(x,z) < d \mu(x),
\end{equation}
\begin{equation}\nonumber
\int_X d \gamma^*(x,z) < d \nu(z).
\end{equation}
Let's set 
\begin{displaymath}
\bigtriangleup m = min\Bigg\{d \mu(x) - \int_Z d \gamma^*(x,z), \; d \nu(z) - \int_X d \gamma^*(x,z)\Bigg\}.
\end{displaymath}
We can increase the mass  until at least one of the above two inequalities holds. The objective will decrease along with the increase of mass. Specifically, while increasing $d \gamma^*(x,z)$ by $\bigtriangleup m$ which still satisfies the partial mass constraints, the objective of the adaptive optimal transport problem will decrease by $-c(x,z) \cdot \bigtriangleup m$. Therefore we obtain a better solution, which contradicts with the premise that $\gamma^*$ is the optimizer. 
\end{proof}

Denote $spt \gamma$ the support of $\gamma$, which refers to the smallest closed subset of $X \times Z$ carrying the full mass of $\gamma$. Define the active regions for the optimizer $\gamma^*$ as
\begin{equation}
X^A = \big\{x \in X \mid \exists z, (x,z) \in spt \gamma^* \big\}
\end{equation}
\begin{equation}
Z^A = \big\{z \in Z \mid \exists x, (x,z) \in spt \gamma^* \big\}
\end{equation}
The inactive regions are denoted as $X^I = X \backslash X^A$ and $Z^I = Z \backslash Z^A$. Denote the complete mass of $\gamma^*$ as $m_{\gamma^*}$. According the definition of active regions, it is straightforward to derive the following theorem regarding active and inactive regions.

\vspace{2mm}
\begin{theorem}[\textbf{Active Regions vs. Inactive Regions}]\label{ar}
There is no mass transferred between inactive regions $X^I \times Z^I$, while the active regions $X^A \times Z^A$ carry the complete mass $m_{\gamma^*}$, i.e.,
\begin{equation}
\gamma^*[X^I \times Z^I] = 0
\end{equation}
\begin{equation}
\gamma^*[X^A \times Z^A] = \gamma^*[X \times Z] = m_{\gamma^*}
\end{equation}
\end{theorem}
The proof is omitted.

\begin{figure}[t]
\centering
\includegraphics[width=7cm]{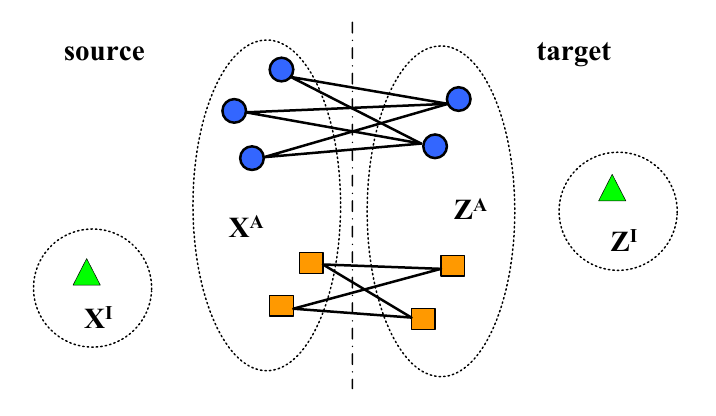}
\vspace{1mm}
\caption{A toy example illustrating the partial distribution alignment via adaptive optimal transport. The line linked two samples represents the mass transport between them. The masses are transferred between the active regions $X^A \cup Z^A$, while there is no transportation of masses between inactive regions $X^I \cup Z^I$. Also, the samples connected with lines form the clusters in active regions, and the isolated samples in inactive regions are likely to be outliers or noises.}
\label{fig:toy}
\end{figure}

Next, we use a toy example to provide an intuitive illustration of adaptive optimal transport. For simplicity, we adopt the discrete setting of adaptive optimal transport here. The source and target domains are denoted by $X = \{x_1, x_2, x_3, x_4, x_5, x_6\}$ and $Z = \{z_1, z_2, z_3, z_4, z_5\}$ respectively. Given the marginals $\mu$ and $\nu$ with uniform probability distribution, and the cost matrix $\mathcal{C}$ as follows
\begin{equation}\nonumber
\mu = 
\renewcommand\arraystretch{1.5}
\begin{bmatrix}
\frac{1}{6} & \frac{1}{6} & \frac{1}{6} & \frac{1}{6} & \frac{1}{6} & \frac{1}{6}
\end{bmatrix},
\quad
\nu = 
\renewcommand\arraystretch{1.5}
\begin{bmatrix}
\frac{1}{5} & \frac{1}{5} & \frac{1}{5} & \frac{1}{5} & \frac{1}{5} 
\end{bmatrix},
\end{equation}
\begin{equation}\nonumber
\mathcal{C} = 
\begin{bmatrix}
-1 & -1 & 1 & 1 & 3   \\
-1 & -1 & 2 & 1 & 1   \\
-1 & -1 & 1 & 1 & 2   \\
2 & 3 & -1 & -1 & 1   \\
1 & 1 & -1 & -1 & 3   \\
1 & 3 & 2 & 1 & 2   \\
\end{bmatrix},
\end{equation}
we can obtain the optimal transport plan $\gamma^*$ and its left and right marginals $\mu_{\gamma^*}$ and $\nu_{\gamma^*}$ as follows
\begin{equation}\nonumber
\gamma^* = 
\renewcommand\arraystretch{1.2}
\begin{bmatrix}
\frac{1}{15} & \frac{1}{15} & 0 & 0 & 0   \\
\frac{1}{15} & \frac{1}{15} & 0 & 0 & 0   \\
\frac{1}{15} & \frac{1}{15} & 0 & 0 & 0   \\
0 & 0 & \frac{1}{12} & \frac{1}{12} & 0   \\
0 & 0 & \frac{1}{12} & \frac{1}{12} & 0   \\
0 & 0 & 0 & 0 & 0   \\
\end{bmatrix},
\end{equation}
\begin{equation}\nonumber
\mu_{\gamma^*} = 
\renewcommand\arraystretch{1.3}
\begin{bmatrix}
\frac{2}{15} & \frac{2}{15} & \frac{2}{15} & \frac{1}{6} & \frac{1}{6} & 0
\end{bmatrix},
\nu_{\gamma^*} = 
\renewcommand\arraystretch{1.3}
\begin{bmatrix}
\frac{1}{5} & \frac{1}{5} & \frac{1}{6} & \frac{1}{6} & 0 
\end{bmatrix}.
\end{equation}
From the optimal transport plan $\gamma^*$, we observe that the active regions $X^A = \{x_1, x_2, x_3, x_4, x_5\}$ and $Z^A = \{z_1, z_2, z_3, z_4\}$ forms two clusters, $\{x_1, x_2, x_3, z_1, z_2\}$ and $\{x_4, x_5, z_3, z_4\}$. The points in inactive regions $X^I = \{x_6\}$ and $Z^I = \{z_5\}$ are likely to be the outliers. This toy example is intuitively illustrated in Figure \ref{fig:toy}. The total mass to be transferred is 
\begin{displaymath}
m_{\gamma^*} = \gamma^*[X^A \times Z^A] = \gamma^*[X \times Z] = \frac{11}{15} < 1.
\end{displaymath}
In this example, although $\mu$ and $\nu$ are probability measures ($\mu[X]=\nu[Z]=1$), the transport plan $\gamma$ is not necessarily to be a probability measure. 
Note that the classical optimal transport must allocate the mass to outliers to meet the full-mass constraint. In contrast, adaptive optimal transport conducts partial distribution alignment adaptively and filters out outliers automatically. Therefore adaptive optimal transport provides a flexible and adaptive solution for distribution alignment. Also, it is worth noting that different cost matrix $\mathcal{C}$ will lead to different optimizer $\gamma^*$, as well as the total mass $m_{\gamma^*}$. Therefore, the optimal transport mass $m_{\gamma^*}$ depends solely on the naive structures of the problem itself.

\subsection{Duality Theory}
Next we derive the duality theory for adaptive optimal transport. We borrow the idea from \cite{Caffarelli2010FreeBI} to reformulate the partial mass transport problem into the complete mass transport problem. However, we overcome the drawback of \cite{Caffarelli2010FreeBI} that needs to specify the fixed budget of mass. In contrast, adaptive optimal transport is able to automatically find the optimal mass to be transferred.   

\vspace{2mm}
\noindent\textbf{Augmented Problem.}
Let's attach an isolated point $\hat\infty$ to $X$ and $Y$, denoted by $\hat{X} = X \cup \{\hat\infty\}$ and $\hat{Z} = Z \cup \{\hat\infty\}$, and extend the cost function
\begin{equation}
\hat{c}(x,z) = \bigg\{ 
\begin{array}{l}
c(x,z)  \qquad if \; x \neq \hat\infty \; and \; z \neq \hat\infty  \\
0		 \qquad \qquad \, otherwise
\end{array}.
\end{equation}
Extend the measures $d\mu(x) = f(x) dx$ and $d\nu(z) = g(z) dz$ to $\hat{X}$ and $\hat{Y}$ by adding a Dirac mass isolated at infinity 
\begin{equation}
\hat\mu = \mu + \|g\|_{L^1} \delta_{\hat\infty}
\end{equation}
\begin{equation}
\hat\nu = \nu + \|f\|_{L^1} \delta_{\hat\infty}
\end{equation}
where $\delta$ is the Dirac function and $\|\cdot\|_{L^1}$ is the $L^1$ norm. A bijection between $\gamma \in \Gamma_{\le}(\mu,\nu)$ and $\hat\gamma \in \Gamma(\hat\mu,\hat\nu)$ is given by
\begin{equation}
\hat\gamma = \gamma + (f-f_\gamma) \otimes \delta_{\hat\infty} + \delta_{\hat\infty} \otimes (g-g_\gamma) + m_{\gamma} \delta_{(\hat\infty,\hat\infty)}
\end{equation}
where $f_\gamma$ and $g_\gamma$ represent the left and right marginals of $\gamma$ respectively, and $\otimes$ is the Kronecker product. Due to mass conservation, we have $\|f_{\gamma}\|_{L^1} = \|g_{\gamma}\|_{L^1}$ and $m_{\gamma} =  \gamma[X \times Z]$.

For the primary problem defined in Equation \ref{pp}, its augmented problem is formulated as
\begin{equation}\label{ap}
\mathop{\min}\limits_{\hat\gamma \in \Gamma(\hat\mu,\hat\nu)} \int_{\hat{X} \times \hat{Z}} \hat{c}(x, z) d \hat\gamma(x,z)
\end{equation}
where the set of admissible transport plans is denoted by
\begin{equation}
\Gamma(\hat\mu,\hat\nu) = \Bigg\{ \hat\gamma \in P(\hat{X} \times \hat{Z}) \hspace{1mm} \bigg| 
\begin{array}{l}
\hat\gamma[A \times \hat{Z}] = \hat\mu[A] \\  
\hat\gamma[\hat{X} \times B] = \hat\nu[B]  
\end{array}
\Bigg\}
\end{equation}
for all Borel subset $A \subset \hat{X}$ and $B \subset \hat{Z}$. \\

\vspace{2mm}
\begin{theorem}[\textbf{Duality of Adaptive Optimal Transport}]\label{duality}
Minimizing the primary adaptive optimal transport problem in Equation \ref{pp} is equivalent to maximizing its dual problem, i.e.,
\begin{equation}\label{pp2dpp}
\begin{aligned}
& \mathop{\min}\limits_{\gamma \in \Gamma_{\le}(\mu,\nu)} \int_{X \times Z} c(x, z) d \gamma(x,z) \\
=  & \mathop{\max}\limits_{\phi(x) + \psi(z) \le c(x, z) \atop \phi(x), \psi(z) \le 0 } \int_{X} \phi(x) d \mu(x) + \int_{Z} \psi(z) d \nu(z)
\end{aligned}
\end{equation}
where $\phi(x) \in C_b(X)$ and $\psi(z) \in C_b(Z)$ are bounded continuous functions. 
\end{theorem}

\vspace{2mm}
\begin{proof}
First, from the bijection between $\gamma$ and $\hat\gamma$, it is easy to see that the primary adaptive optimal transport problem is equivalent to the augmented problem, i.e.,
\begin{equation}\label{pp2ap}
\begin{aligned}
& \mathop{\min}\limits_{\gamma \in \Gamma_{\le}(\mu,\nu)} \int_{X \times Z} c(x, z) d \gamma(x,z) \\
=  & \mathop{\min}\limits_{\hat\gamma \in \Gamma(\hat\mu,\hat\nu)} \int_{\hat{X} \times \hat{Z}} \hat{c}(x, z) d \hat\gamma(x,z)
\end{aligned}
\end{equation}

Second, for the full-mass optimal transport, by Kantorovich duality \cite{KantorovitchOnTT}, we can prove the equivalence between the augmented problem and its Kantorovich dual problem, i.e.,
\begin{equation}\label{ap2dap}
\begin{aligned}
& \mathop{\min}\limits_{\hat\gamma \in \Gamma(\hat\mu,\hat\nu)} \int_{\hat{X} \times \hat{Z}} \hat{c}(x, z) d \hat\gamma(x,z) \\
=  & \mathop{\max}\limits_{\hat\phi(x) + \hat\psi(z) \le \hat{c}(x, z)} \int_{\hat{X}} \hat\phi(x) d \hat\mu(x) + \int_{\hat{Z}} \hat\psi(z) d \hat\nu(z)
\end{aligned}
\end{equation}
where $\hat\phi(x) \in C_b(\hat{X})$ and $\hat\psi(z) \in C_b(\hat{Z})$ are bounded continuous functions.
It is similar to the proof of standard Kantorovich duality (please refer to Section 1.1.5 of \cite{Villani2003TopicsIO} for details). The basic idea is to remove the constraints by using the generalized Lagrange multiplier method. Note that as shown in the right-hand of Equation \ref{ap2dap}, the duality of the augmented problem requires that $\hat\phi(x) + \hat\psi(z) \le \hat{c}(x, z)$, i.e.,  $\hat\phi(x) + \hat\psi(z) \le c(x, z)$ if $x \neq \hat\infty$ and $z \neq \hat\infty$ as usually, and $\hat\phi(x) + \hat\psi(z) \le 0$ otherwise to accommodate the isolated point $\infty$.

Third, it remains to prove the equivalence between the duality of the primary optimal transport problem and the duality of its augmented problem, i.e.,
\begin{equation}\label{dap2dpp}
\begin{aligned}
& \mathop{\max}\limits_{\hat\phi(x) + \hat\psi(z) \le \hat{c}(x, z)} \int_{\hat{X}} \hat\phi(x) d \hat\mu(x) + \int_{\hat{Z}} \hat\psi(z) d \hat\nu(z) \\
=  & \mathop{\max}\limits_{\phi(x) + \psi(z) \le c(x, z) \atop \phi(x), \psi(z) \le 0 } \int_{X} \phi(x) d \mu(x) + \int_{Z} \psi(z) d \nu(z)
\end{aligned}
\end{equation}
where $\phi(x) \in C_b(X)$ and $\psi(z) \in C_b(Z)$ are bounded continuous functions. 

For Equation \ref{dap2dpp}, any competitors $(\phi,\psi)$ in the right-hand can be extended to $\hat{X}$ and $\hat{Y}$ by taking $\phi(\hat\infty)=\psi(\hat\infty)=0$. This extension still satisfied the constraints of the left-hand side. The maximization of the left-hand side of Equation \ref{dap2dpp} over larger class of competitors can only dominate the maximization of its right-hand side.

For the left-hand side of Equation \ref{dap2dpp}, since
\begin{equation}\nonumber
\max  \int_{\hat{X}} (\hat\phi+k) d \hat\mu + \int_{\hat{Z}} (\hat\psi-k) d \hat\nu = \max  \int_{\hat{X}} \hat\phi d \hat\mu + \int_{\hat{Z}} \hat\psi d \hat\nu
\end{equation}
for any $k \in \mathbb{R}$, we are free to assume $\hat\phi(\hat\infty) = 0$. By the constraint $\hat\phi(\hat\infty) + \hat\psi(z) \le \hat{c}(\hat\infty,z) = 0$ for any $z \in \hat{Z}$, we obtain $\hat\psi(z) \le 0$ throughout $z \in \hat{Z}$. At $z = \hat\infty$, the only constraint is that 
\begin{equation}\nonumber
\hat\psi(\hat\infty) \le \mathop{\inf}\limits_{x \in \hat{X}} \hat{c}(x, \hat\infty) - \hat\phi(x) = \mathop{\inf}\limits_{x \in \hat{X}} -\hat\phi(x) = -\phi_{max}
\end{equation}
and the equality can be assumed to hold for the optimizer $(\hat\phi, \hat\psi)$ of the left-hand side of Equation \ref{dap2dpp}. Thus
\begin{equation}\nonumber
\begin{aligned}
   & \int_{\hat{X}} \hat\phi d \hat\mu + \int_{\hat{Z}} \hat\psi d \hat\nu  \\
= & \int_{X} \hat\phi d \mu + \hat\phi(\hat\infty) \|g\|_{L^1} + \int_{Z} \hat\psi d \nu + \hat\psi(\hat\infty) \|f\|_{L^1} \\
= & \int_{Z} \hat\psi d \nu + \int_{X} \hat\phi d \mu - \phi_{max} \|f\|_{L^1}
\end{aligned}
\end{equation}
The sum of the last two terms is not positive since $\phi_{max} \ge \hat\phi(\hat\infty) = 0$. Replacing $\phi$ by $min\{\phi, 0\}$ pointwise always increases the above objective since $\phi - \phi_{max} \le min\{\phi, 0\}$, and makes it easier to satisfy the constraints of the right-hand side of Equation \ref{dap2dpp}. Therefore, we conclude $\phi_{max} = 0$. Thus
\begin{equation}\nonumber
\int_{\hat{X}} \hat\phi d \hat\mu + \int_{\hat{Z}} \hat\psi d \hat\nu = \int_{X} \hat\phi d \mu + \int_{Z} \hat\psi d \nu
\end{equation}
By now, we have shown that $\hat\phi(x) \le 0 (\forall x \in \hat{X})$ and $\hat\psi(z) \le 0 (\forall z \in \hat{Z})$. That is to say, the restriction $(\phi, \psi)$ of $(\hat\phi, \hat\psi)$ to $X \times Z$ now satisfies the constraint of the right-hand side of Equation \ref{dap2dpp}. Therefore, for Equation \ref{dap2dpp}, the maximization of the right-hand side dominates the maximization of the left-hand size. Hence the two maximum values coincide, which completes the proof.
\end{proof}

The $c$-transform and $\bar{c}$-transform is defined as
\begin{equation}
\phi^c(z) = \mathop{\inf}\limits_{x \in X} c(x,z) - \phi(x)
\end{equation}
\begin{equation}
\psi^{\bar{c}}(x) = \mathop{\inf}\limits_{z \in Z} c(x,z) - \psi(z)
\end{equation}
Using $c$-transform and $\bar{c}$-transform, one can reformulate the duality of the adaptive optimal transport problem over two potentials as an convex program over a single potential
\begin{equation}\label{dpp2fg}
\begin{aligned}
& \mathop{\max}\limits_{\phi(x) + \psi(z) \le c(x, z) \atop \phi(x), \psi(z) \le 0 } \int_{X} \phi(x) d \mu(x) + \int_{Z} \psi(z) d \nu(z)  \\
=  & \mathop{\max}\limits_{\phi(x), \phi^c(z) \le 0 } \int_{X} \phi(x) d \mu(x) + \int_{Z} \phi^c(z) d \nu(z)	   \\
=  & \mathop{\max}\limits_{\psi(z), \psi^{\bar{c}}(x) \le 0 } \int_{X} \psi^{\bar{c}}(x) d \mu(x) + \int_{Z} \psi(z) d \nu(z)
\end{aligned}
\end{equation}

According to Theorem \ref{duality}, we can see that the duality of adaptive optimal transport is similar to the Kantorovich duality of the full-mass OT problem, except for the additional constraints $\phi(x) \le 0, \psi(z) \le 0$. Therefore, the off-the-shelf algorithms for Kantorovich duality can be adapted for solving the adaptive optimal transport problem by imposing the additional constraints $\phi(x) \le 0, \psi(z) \le 0$, e.g., replacing $\phi$ by $min\{\phi, 0\}$, and $\psi$ by $min\{\psi, 0\}$ pointwise. \\

\section{Experiments}\label{sec5}
In the experiment section, we mainly focus on answering three questions:
\begin{itemize}
\item How are the probability masses adaptively transferred from source domain to target domain via adaptive optimal transport? 
\item How is the robustness of adaptive optimal transport in the scenarios of partial matching?
\item How well does the adaptive optimal transport method behave in comparison with the state-of-the-art algorithms, especially the classical optimal transport and partial optimal transport approaches?
\end{itemize}

\begin{figure*}[htbp]
\centering
\subfloat[$\epsilon=0.1$]{
\begin{minipage}[t]{0.46\textwidth}
\centering
\includegraphics[width=\columnwidth]{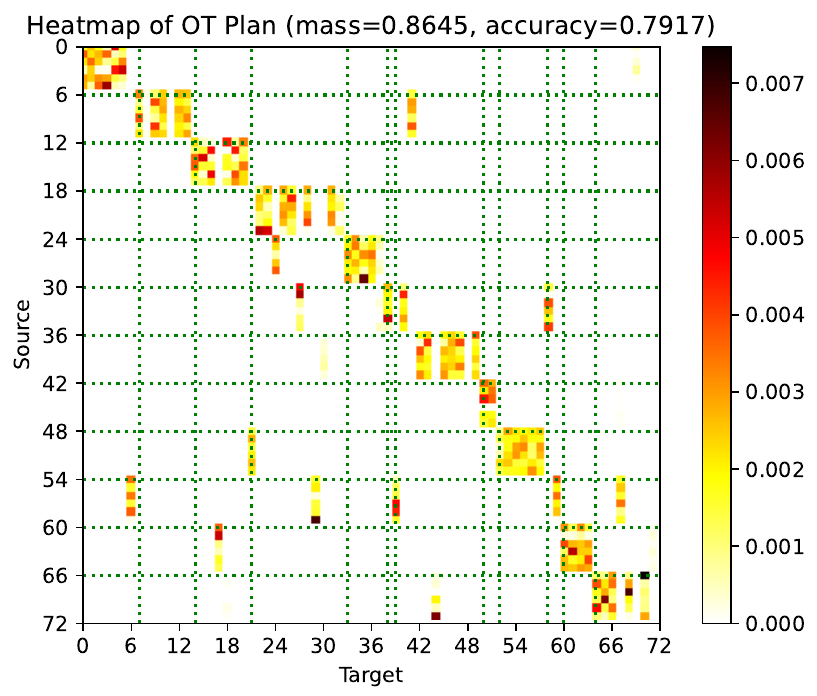}
\label{fig:ot2}
\vspace{-3mm}
\end{minipage}
}\hspace{3mm}
\subfloat[$\epsilon=0$]{
\begin{minipage}[t]{0.46\textwidth}
\centering
\includegraphics[width=\columnwidth]{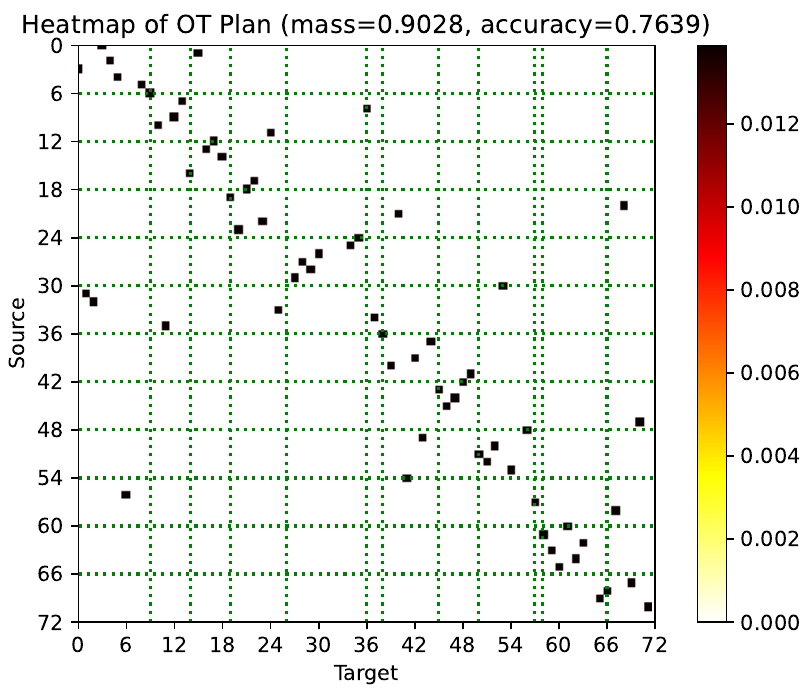}
\label{fig:ot1}
\vspace{-3mm}
\end{minipage}
}
\vspace{2mm}
\caption{Heatmap of optimal transport plan illustrating the mass allocation mechanism of adaptive optimal transport. The $72 \times 72$ transport plan matrix is partitioned into $12 \times 12$ blocks. Most of the masses are allocated along the diagonal blocks, aligning labels between source domain and target domain.}
\label{fig:hm}
\end{figure*}

\subsection{Experiment Setup}
For a fair comparison. we basically follow the settings of JUMBOT~\cite{fatras2021unbalanced} and m-POT~\cite{NguyenNV0H22} for the experiment setups. 

\textbf{Datatsets.} We use three domain adaptation benchmark datasets in the experiments. \textbf{VisDA}~\cite{peng2017visda} is a large-scale dataset for unsupervised domain adaptation. It contains 152,397 synthetic images as the source domain and 55,388 real-world images as the target domain. The two domains share 12 object categories. Following the common setting~\cite{fatras2021unbalanced,NguyenNV0H22}, we evaluate all methods on VisDA validation set. \textbf{Office-Home}~\cite{venkateswara2017deep} contains 15,500 images from four domains: Artistic images (A), ClipArt (C), Product images (P) and Real-World (R). For each domain, it consists of 65 object categories that are common in home and office scenarios. All methods are evaluated on 12 adaptation tasks. \textbf{Office-31}~\cite{saenko2010adapting} consists of 4652 images from 31 categories, collected from three domains including Amazon (2817 images), Webcam (795 images) and DSLR (498 images), respectively. There are totally 6 adaptation tasks for evaluation. 

\textbf{Networks.} Note that using some more advanced backbones may lead to better performance. But for the fair comparison, we adopt ResNet-50~\cite{he2016identity} as backbone, which is the same with \cite{fatras2021unbalanced, NguyenNV0H22}. The ResNet-50 pretrained on ImageNet is used as feature extractor and one fully connected (FC) layer is used as classifier for all three datasets. 
 
\textbf{Sampling.} Similar to the previous work \cite{fatras2021unbalanced,NguyenNV0H22}, we adopt the stratified sampling to select a mini-batch of source samples so that each class has the same number of samples. The random sampling is used on target domain since labels are unavailable in training. 

\textbf{Data Augmentation.} Following \cite{fatras2021unbalanced, NguyenNV0H22},  we use the same data pre-processing for all three datasets. The images are first resized into $256\times 256$ and then randomly cropped with size of $224\times 224$. Random translation/mirror and normalization are also applied for training. For testing, we adopt the ten-crop technique~\cite{fatras2021unbalanced, NguyenNV0H22} for robust results. Note that these settings are commonly used and the same as previous works~\cite{fatras2021unbalanced, NguyenNV0H22} for fair comparison.

\textbf{Training Details.} Following the settings of ~\cite{fatras2021unbalanced, NguyenNV0H22}, we adopt SGD optimizer with 0.9 momentum and $5e^{-4}$ weight decay for training, and the learning rates are set with the same strategy as~\cite{ganin2016domain}. Note that the learning rate of the classifier is set to be 10 times that of the extractor as the classifier is trained from scratch. 

\textbf{Hyper-parameters.} For all three datasets, the weight of feature-wise cost $\alpha$ is set to 0.01. The weight of label-wise cost $\beta$ is set to 1.8, 6, and 5 for VisDA, Office-Home, and Office-31 respectively. The entropy-regularized coefficient $\epsilon$ is set to 0.1, 1, and 1 for VisDA, Office-Home, and Office-31 respectively. The batch size is set to 72, 65, and 62 for VisDA, Office-Home, and Office-31, respectively. The experiments are trained for 2000, 5000, and 5000 iterations for VisDA, Office-Home, and Office-31, respectively.

\subsection{Adaptive Mass Transport}
The first issue is how the probability masses are adaptively transported from source domain to target domain via adaptive optimal transport. Therefore, we visualize the heatmaps of the transport plans to provide an intuitive illustration of the adaptive mechanism of mass allocation. 

Figure \ref{fig:hm} plots the heatmaps of optimal transport plans in a mini-batch for the VisDA dataset. Each row (or column) corresponds to one source (or target) sample. The source (or target) samples are reordered into clusters by the order of labels. Note that the batch size is set to $b=72$ and the number of classes is 12. Therefore, the $72 \times 72$ transport plan matrix is partitioned into $12 \times 12$ blocks. Because the random sampling strategy is adopted in the target domain, the numbers of samples for the target labels are uneven. As expected, the masses are almost allocated along the diagonal blocks, aligning labels between source and target domains. It suggests that the optimal transport in AOT is class-aware. Figure \ref{fig:hm} also shows the heatmaps with $\epsilon=0.1$ and $\epsilon=0$ in the left and right panels respectively. It intuitively demonstrated that the transport plan becomes sparser as the entropy-regularized coefficient $\epsilon$ increases. As shown on the top of the figures, the total masses transported from source domain to target domain are 0.8645 and 0.9028 for  $\epsilon=0.1$ and $\epsilon=0$, respectively. It is worth noting that the total masses are self-determined by AOT, which could be roughly regarded as the relatedness between the source and target domains. In contrast, partial optimal transport \cite{Caffarelli2010FreeBI,Figalli2010TheOP} has to pre-define the fixed budget of masses, which remains a challenging issue. 

In summary, the heatmap intuitively verifies that AOT transports masses from source domain to target domain in an adaptive and class-aware way. Our method can automatically learn the optimal fraction of masses to be transported, leading to an elegant solution for partial distribution alignment.

\begin{figure*}[htbp]
\centering
\subfloat[Missing Label is `9']{
\begin{minipage}[t]{0.46\textwidth}
\centering
\includegraphics[width=\columnwidth]{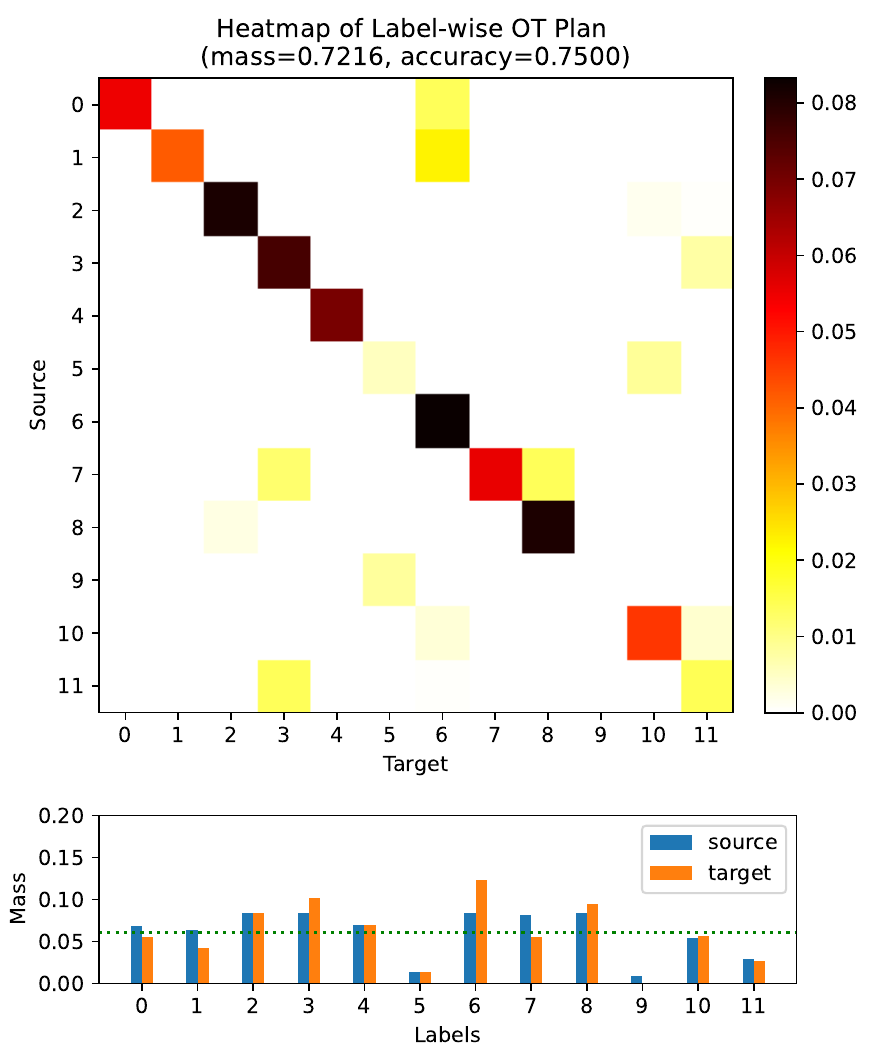}
\label{fig:ot3}
\vspace{-3mm}
\end{minipage}
}\hspace{3mm}
\subfloat[Missing Label is `0']{
\begin{minipage}[t]{0.46\textwidth}
\centering
\includegraphics[width=\columnwidth]{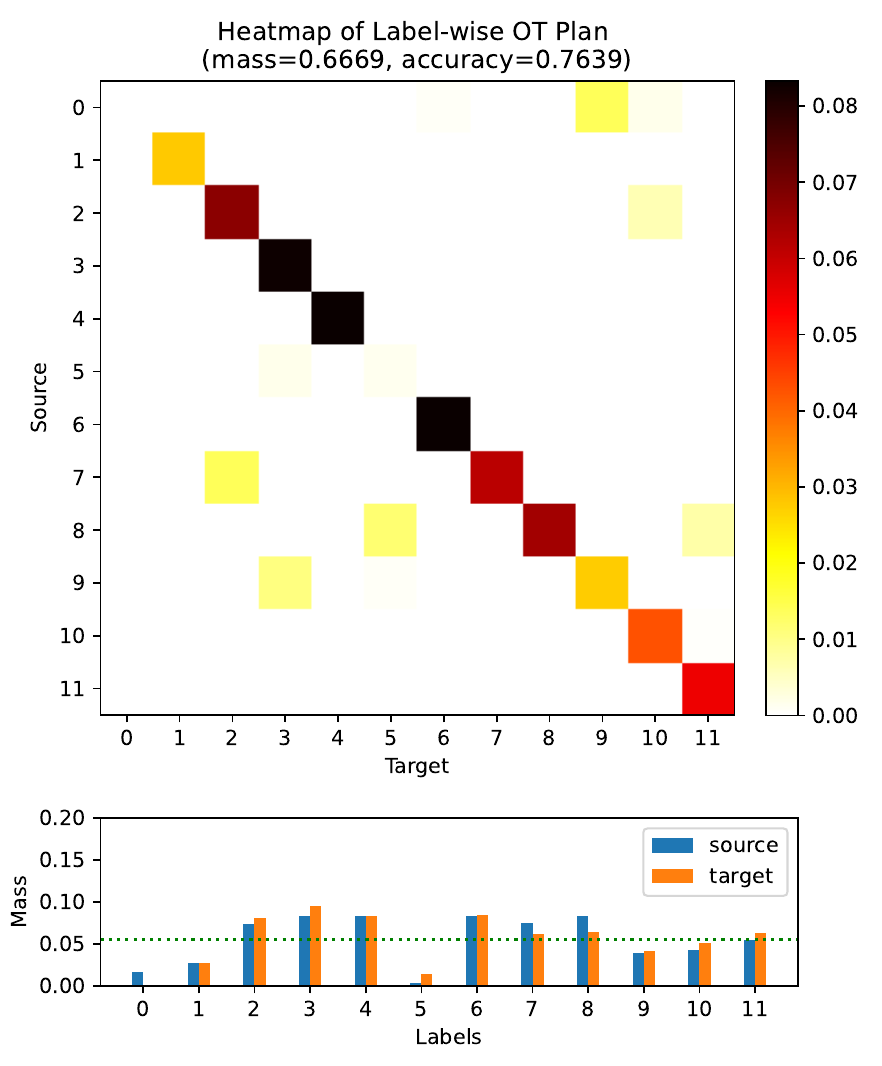}
\label{fig:ot4}
\vspace{-3mm}
\end{minipage}
}
\vspace{2mm}
\caption{Heatmap of label-wise transport plan in the case of partial mapping (top panels). The masses are aggregated by the labels. The histograms plot the label-wise marginal distributions for the source and target domains respectively (bottom panels). }
\label{fig:hm2}
\end{figure*}

\subsection{Robustness of Partial Mapping}
The second targeting issue is the robustness of adaptive optimal transport in the scenarios of partial matching.

Partial matching is rather common in real applications. For example, because the random sampling strategy is adopted in the target domain, it is possible that some target labels are missing in a mini-batch. There is no one-to-one correspondence between the source labels and the target labels. It prefers partial matching to complete matching in these situations. Therefore, we take this scenario as an example to investigate the robustness of adaptive optimal transport for partial matching. 

Similar to Figure \ref{fig:hm} that shows the sample-wise transport plan, Figure \ref{fig:hm2} plots the heatmaps of the label-wise transport plan. The transport plan matrix is reorganized by the labels and then partitioned into $12 \times 12$ blocks, each of which aggregates the masses of sample pairs with the corresponding source and target labels. The histograms under the heatmaps display the label-wise marginal distributions of the OT plan. The green horizontal line serves as the mean of marginal distributions. The left and right panels of Figure \ref{fig:hm2} indicate the cases with missing target labels, `9' and `0', respectively. It is observed from both histograms that the source label-wise marginals are approximate to the target ones. Again it verified that AOT adaptively transports masses in accordance with labels. Furthermore, as shown in the left panel, when the target label `9' is absent, the source marginal for label `9' is far below the average line. It suggests that AOT is able to automatically reduce the mass for the source label `9' to transmit, in response to the fact that the corresponding target label is missing. On the contrary, classical optimal transport \cite{KantorovitchOnTT} with full-mass constraint will keep the budget mass unchanged regardless of the missing labels. The similar phenomenon is observed in the right panel when the target label `0' is absent. 

The results demonstrate the adaptiveness of AOT for partial matching, giving the clear evidences to support our theoretical analysis on adaptive optimal transport.

\begin{table}[htbp]
\centering
\caption{Classification accuracy on VisDA (ResNet-50), where (*) and (\#) denote the results quoted from m-POT~\cite{NguyenNV0H22} and JUMBOT~\cite{fatras2021unbalanced}, respectively.}
\begin{tabular}{cc}
\toprule
Method & Accuracy \\
\midrule
DANN$^{(*)}$  & 67.63$\pm$0.34  \\
CDAN$^{(\#)}$ & 70.10 \\
ALDA$^{(*)}$ & 71.22$\pm$0.12 \\
ROT$^{(\#)}$ & 66.30\\
DeepJDOT$^{(\#)}$ & 68.00 \\
JUMBOT$^{(\#)}$ & 72.50 \\
m-POT$^{(*)}$ & \underline{73.59}$\pm$0.15 \\
\textbf{AOT (ours)} & \textbf{76.68}$\pm$0.19\\
\bottomrule
\end{tabular}
\label{tab:visda}
\end{table}

\begin{table*}[htbp]
\centering
\caption{Classification accuracy on Office-Home (ResNet-50), where (*) denotes the results quoted from m-POT~\cite{NguyenNV0H22}.}
\setlength{\tabcolsep}{0.4mm}
{
\begin{tabular}{cccccccccccccc}
\toprule
Method & A-C & A-P & A-R & C-A & C-P & C-R & P-A & P-C & P-R & R-A & R-C & R-P & Avg \\
\midrule
ResNet-50${(*)}$ & 34.90 & 50.00 & 58.00 & 37.40 & 41.90 & 46.20 & 38.50 & 31.20 & 60.40 & 53.90 & 41.20 & 59.90 & 46.10\\
DANN${(*)}$ & 47.92 & 67.08 & 74.85 & 53.80 & 63.47 & 66.42 &  52.99 & 44.35 & 74.43 & 65.53 & 52.96 & 79.41 & 61.93\\
CDAN${(*)}$ & 52.50 & 71.40 & 76.10 & 59.70 & 69.90 & 71.50  &  58.70 & 50.30 & 77.50 & 70.50 & 57.90 & 83.50 &66.60\\
ALDA${(*)}$ & 54.04 & \underline{74.89} & 77.14 & 61.37 & 70.62 & 72.75 &  60.32 & 51.03 & 76.66 & 67.90 & 55.94 & 81.87 & 67.04\\
ROT${(*)}$ & 47.20 & 70.80 & 76.40 & 58.60 & 68.10 & 70.20  & 56.50 & 45.00 & 75.80 & 69.40 & 52.10 & 80.60 & 64.30\\
DeepJDOT${(*)}$ & 51.75 & 70.01 & 75.59 & 59.60 & 66.46 & 70.07  &  57.60 & 47.88 & 75.29 & 66.82 & 55.71 & 78.11 & 64.59\\
JUMBOT${(*)}$ & 54.99 & 74.45 & \underline{80.78} & 65.66 & \underline{74.93} & 74.91  &  \underline{64.70} & \underline{53.42} & 80.01 & \underline{74.58} & \underline{59.88} & 83.73 & 70.17\\
m-POT${(*)}$ & \underline{55.65} & 73.80 & 80.76 & \underline{66.34} & 74.88 & \underline{76.16}  &  64.46 & 53.38 & \underline{80.60} & 74.55 & 59.71 & \underline{83.81}  & \underline{70.34} \\
{\textbf{AOT(ours)}} & \textbf{56.94}$\pm$0.1  & \textbf{78.31}$\pm$0.1  & \textbf{82.97}$\pm$0.1  & \textbf{71.12}$\pm$0.2   & \textbf{74.68}$\pm$0.1  & \textbf{78.79}$\pm$0.2  & \textbf{66.50}$\pm$0.3  & \textbf{54.80}$\pm$0.2 & \textbf{82.44}$\pm$0.1  & \textbf{75.48}$\pm$0.1  & \textbf{60.18}$\pm$0.2  & \textbf{84.72}$\pm$0.1  & \textbf{72.24} \\
\bottomrule
\end{tabular}
}
\label{tab:oh}

\vspace{10mm}
\caption{Classification accuracy on Office-31 (ResNet-50). The results reproduced from the official codes of m-POT~\cite{NguyenNV0H22} are denoted with ($\circ$).}
\setlength{\tabcolsep}{1mm}
{
\begin{tabular}{cccccccc}
\toprule
Method & A$\rightarrow$W & D$\rightarrow$W & W$\rightarrow$D &  A$\rightarrow$D & D$\rightarrow$A & W$\rightarrow$A  & Avg\\
\midrule
ResNet-50 & 68.4 & 96.7 & 99.3 & 68.9 & 62.5 & 60.7 & 76.1\\
DANN & 82.0 &  96.9 & 99.1 & 79.7 & 68.2 & 67.4 & 82.2\\
CDAN & 93.1 & \underline{98.6} & \textbf{100.0} & 92.9 & 71.0 & 69.3 & 87.5  \\
ALDA & \underline{95.6} & 97.7 & \textbf{100.0} & \textbf{94.0} & 72.2 & 72.5 & \underline{88.7}\\
CaCo & 89.7 & 98.4 & \textbf{100.0} & 91.7 & \underline{73.1} & \underline{72.8} & 87.6\\
DeepJDOT ($\circ$)  & 87.8$\pm$0.2 & 97.9$\pm$0.3 & 99.8$\pm$0.1 & 88.7$\pm$0.1 & 70.8$\pm$0.3 & 71.3$\pm$0.2 & 86.1\\
JUMBOT ($\circ$) & 91.5$\pm$0.4 & 98.5$\pm$0.2 & \textbf{100.0}$\pm$0   & 89.4$\pm$0.3 & 68.8$\pm$0.3 & 70.2$\pm$0.2 & 86.4\\
m-POT ($\circ$) & 93.7$\pm$0.3 & \textbf{99.1}$\pm$0.1 & \textbf{100.0}$\pm$0 & \underline{93.3}$\pm$0.2 & 70.9$\pm$0.4 & 72.5$\pm$0.1 & 88.3\\
\textbf{AOT(ours)} & \textbf{95.5}$\pm$0.2 & 98.9$\pm$0.1 & \textbf{100.0}$\pm$0 & \textbf{95.7}$\pm$0.3 & \textbf{77.0}$\pm$0.2 & \textbf{78.7}$\pm$0.1 & \textbf{90.9}\\
\bottomrule
\end{tabular}
}
\label{tab:o31}
\end{table*}

\subsection{Performance Comparison}
The third question we aim to answer is how well the adaptive optimal transport method performs against the state-of-the-art algorithms. We are especially interested in performance comparison between adaptive optimal transport with adaptive-mass preservation and the classical optimal transport with full-mass or fixed-mass preservation.

We compare our method with a variety of unsupervised domain adaptation algorithms including: 1) OT-based methods such as ROT~\cite{BalajiCF20}, DeepJDOT~\cite{damodaran2018deepjdot},  JUMBOT~\cite{fatras2021unbalanced}, and m-POT~\cite{NguyenNV0H22}; and 2) Non-OT-based methods such as 
DANN~\cite{ganin2016domain}, CDAN~\cite{long2018conditional}, ALDA~\cite{chen2020adversarial}, and CaCo~\cite{huang2022category}. 
For a fair comparison, the backbones of all the methods are based on the deep neural network ResNet-50~\cite{he2016identity} pretrained on ImageNet. We conduct each experiment three times and report the average Accuracy score (in \%) and standard deviation. The accuracies of the comparison methods are reproduced, or quoted from JUMBOT~\cite{fatras2021unbalanced}, m-POT~\cite{NguyenNV0H22} or their own papers unless otherwise stated. The standard deviations for the comparison methods are shown whenever available in their papers.  

Tables \ref{tab:visda}, \ref{tab:oh}, and \ref{tab:o31} show that the proposed AOT method significantly outperforms the comparison baselines on three benchmark datasets. The bold and underlined accuracies represent the best and the second best performances, respectively. On the large-scale VisDA dataset which is one or two order of magnitude higher than the other two datasets, AOT beats the runner-up method m-POT by a clear margin. On Office-Home dataset, AOT performs consistently better than the comparison methods on all 12 domain adaptation tasks. For Office-31 dataset, AOT also achieves higher accuracies in 5 out of 6 adaptation scenarios. 

We take a closer look at the unsupervised domain adaptation methods based on optimal transport. DeepJDOT~\cite{damodaran2018deepjdot} adopted Kantorovich optimal transport to align both feature and label distributions. JUMBOT~\cite{fatras2021unbalanced} and m-POT~\cite{NguyenNV0H22} followed DeepJDOT to align the joint distributions. Therefore, all of DeepJDOT~\cite{damodaran2018deepjdot}, JUMBOT~\cite{fatras2021unbalanced} and m-POT~\cite{NguyenNV0H22} adopted the same cost function 
\begin{displaymath}
c(x,z) = \alpha \big\|x - z\big\|_2^2 - \beta p^T(x) \cdot \log q(z).
\end{displaymath}
which is non-negative. By contrast, we use a different cost function
\begin{displaymath}
c(x,z) = \alpha \big\|x - z\big\|_2^2 - \beta p^T(x) \cdot q(z).
\end{displaymath}
which could be positive or negative. The rationale is that it not only considers the similarity in both feature and label spaces, but also allows for adaptive mass transport in our AOT model. It is also worth noting that AOT has not introduced any new hyperparameters in the cost function. The primary differences among the above methods lie in the optimal transport methods. DeepJDOT~\cite{damodaran2018deepjdot} relied on Kantorovich optimal transport, while JUMBOT~\cite{fatras2021unbalanced} and m-POT~\cite{NguyenNV0H22} used unbalanced optimal transport \cite{Liero2017OptimalEP} and partial optimal transport \cite{Caffarelli2010FreeBI,Figalli2010TheOP} respectively.

JUMBOT and m-POT performed better than DeepJDOT, indicating that both unbalanced optimal transport and partial optimal transport could alleviate the influence of undesired coupling between samples and overcome the limitations of Kantorovich optimal transport to some extent. Tables \ref{tab:visda}, \ref{tab:oh}, and \ref{tab:o31} show that AOT consistently outperforms the above OT based methods including DeepJDOT~\cite{damodaran2018deepjdot}, JUMBOT~\cite{fatras2021unbalanced} and m-POT~\cite{NguyenNV0H22}. It verifies the effectiveness of adaptive optimal transport. By relaxing the full-mass or fixed-mass constraints, AOT exhibits a great flexibility in accommodating the noises, outliers and distribution shifts, leading to better performance. The strength is that AOT relies on the intrinsic structure of the problem itself to transport suitable masses adaptively. As a novel member in the family of optimal transport, AOT provides a principled framework for partial distribution alignment.

\begin{figure}[htbp]
\centering
\includegraphics[width=0.8\columnwidth]{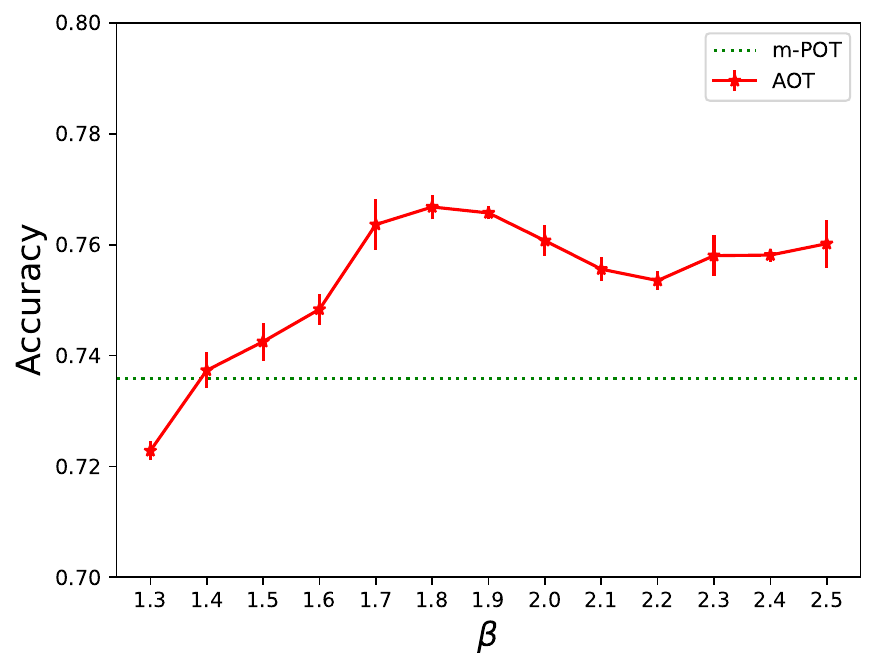}
\caption{Sensitivity analysis on $\beta$.}
\label{fig:para1}
\end{figure}

\begin{figure}[htbp]
\centering
\includegraphics[width=0.8\columnwidth]{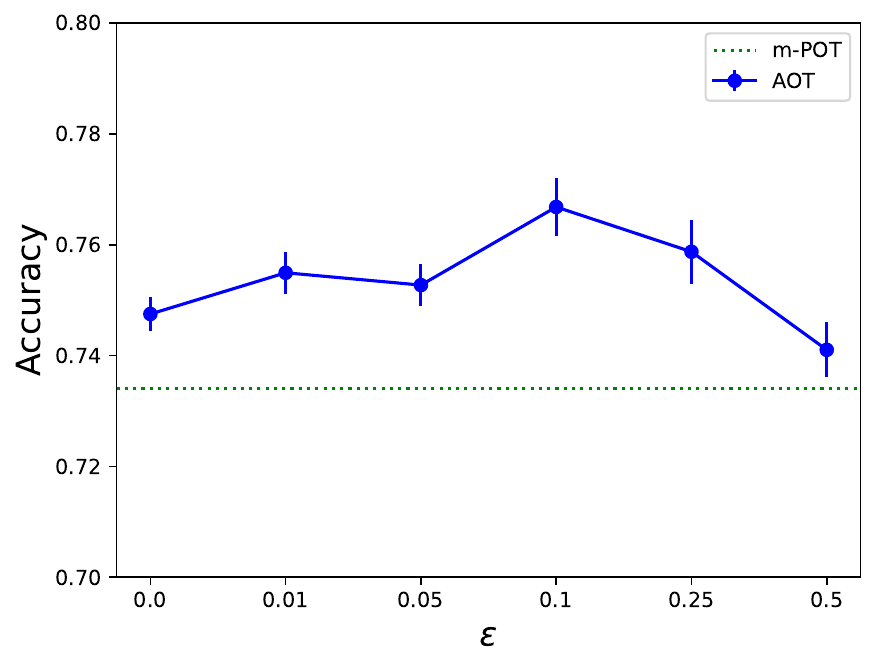}
\caption{Sensitivity analysis on $\epsilon$.}
\label{fig:para2}
\end{figure}

\subsection{Hyperparameter Sensitivity}
To further investigate the robustness of the proposed method for domain adaptation, we report the sensitivity analysis with $\beta$ and $\epsilon$ on VisDA dataset. Figures \ref{fig:para1} and \ref{fig:para2} show the average accuracy and standard deviation varying with the parameters. For comparison, the performance of m-POT~\cite{NguyenNV0H22} is also plotted as base line. Note that $\beta$ is the weight to control the impact of label-wise cost. It is observed from Figure \ref{fig:para1} that the accuracy of AOT reaches maximum around $\beta=1.8$, then drops slightly when $\beta$ becomes larger. Nevertheless when $\beta$ varies in a wide spectrum (e.g., 1.4 to 2.5), AOT beats the second best method m-POT~\cite{NguyenNV0H22}. The entropy-regularized coefficient $\epsilon$ is to control the sparsity of transport plan. Figure \ref{fig:para2} shows that entropy-regularized term improves the performance when $\epsilon$ increases from 0 to 0.1. The accuracy of AOT reaches its maximum around $\epsilon=0.1$. However, when the transport plan becomes more sparse with a larger $\epsilon$, the accuracy falls.

\section{Conclusion}\label{sec6}
We propose adaptive optimal transport to enrich the toolbox of optimal transport. The mechanism of adaptive mass allocation is theoretically exploited, and the effectiveness of adaptive optimal transport is empirically verified on various domain adaptation benchmarks. Due to the ubiquity of noises, outliers, and distribution shifts, a variety of open-world artificial intelligence applications can opt for adaptive optimal transport whenever partial distribution alignment is preferred. As a fundamental tool for partial distribution alignment, we believe that adaptive optimal transport opens the pathway to unlock problems in many areas beyond artificial intelligence. In the future, we will explore the applications of adaptive optimal transport in biomedical domain, such as understanding cell perturbation responses to treatments.

\bibliographystyle{IEEEtran}

\end{document}